\documentclass[conference]{IEEEtran}
\IEEEoverridecommandlockouts
\usepackage{cite}
\usepackage{amsmath,amssymb,amsfonts}
\usepackage{graphicx}
\usepackage{textcomp}
\usepackage{xcolor}
\usepackage{hyperref}
\usepackage[numbers]{natbib}

\usepackage{amsmath}
\usepackage{wrapfig}
\usepackage{algorithm}
\usepackage[noend]{algorithmic}
\usepackage{subcaption}
\usepackage{booktabs}
\usepackage{soul}
\usepackage{multirow}
\usepackage{amsthm}
\theoremstyle{definition}

\usepackage{bm}

\newenvironment{algoblue}{%
   \setlength{\parindent}{0pt}
   \itshape
   \color{blue}
}{}

\newcommand{\eMCTSu}{Elastic MCTS$_u$}

\newcommand{\MCTSu}{MCTS$_u$}
\newcommand{\rMCTSu}{RG MCTS$_u$ }
\newcommand{\rb}{Rule-based }
\newcommand{\SCSA}{SCSA }

\IEEEoverridecommandlockouts
\IEEEpubid{\makebox[\columnwidth]{ 979-8-3503-5067-8/24/\$31.00~\copyright2024 IEEE \hfill} 
\hspace{\columnsep}\makebox[\columnwidth]{ }}

\def\BibTeX{{\rm B\kern-.05em{\sc i\kern-.025em b}\kern-.08em
    T\kern-.1667em\lower.7ex\hbox{E}\kern-.125emX}}
\begin{document}

\title{Strategy Game-Playing with\\
Size-Constrained State Abstraction
}

\author{\IEEEauthorblockN{Linjie Xu, Diego Perez-Liebana}
\IEEEauthorblockA{\textit{School of EECS} \\
\textit{Queen Mary University of London, London, UK} \\
{\{linjie.xu, diego.perez\}@qmul.ac.uk}}
\and
\IEEEauthorblockN{Alexander Dockhorn}
\IEEEauthorblockA{\textit{Faculty of EECS} \\
\textit{Leibniz University Hannover, Hannover, Germany} \\
dockhorn@tnt.uni-hannover.de}
}

\maketitle
\IEEEpubidadjcol

\begin{abstract}
Playing strategy games is a challenging problem for artificial intelligence (AI). One of the major challenges is the large search space due to a diverse set of game components. In recent works, state abstraction has been applied to search-based game AI and has brought significant performance improvements. State abstraction techniques rely on reducing the search space, e.g., by aggregating similar states. However, the application of these abstractions is hindered because the quality of an abstraction is difficult to evaluate. Previous works hence abandon the abstraction in the middle of the search to not bias the search to a local optimum. This mechanism introduces a hyper-parameter to decide the time to abandon the current state abstraction. In this work, we propose a size-constrained state abstraction (SCSA), an approach that limits the maximum number of nodes being grouped together. We found that with SCSA, the abstraction is not required to be abandoned. Our empirical results on $3$ strategy games show that the SCSA agent outperforms the previous methods and yields robust performance over different games. Codes are opensourced at \url{https://github.com/GAIGResearch/Stratega}.
\end{abstract}


\section{Introduction}
Strategy games have helped advance the development of Artificial Intelligence (AI) to achieve significant progress in competing with human players~\citep{vinyals2019grandmaster, berner2019dota}, AI-AI cooperation~\citep{hu2020other, cui2021k, yu2022the} and human-AI cooperation~\citep{hu2020other, hu2021off, strouse2021collaborating, Bakhtin2022}. Most of this progress depends on deep reinforcement learning (DRL). However, DRL agents have their neural networks trained and tuned for a specific game, making it difficult to apply these agents to other game variants. In contrast, search-based algorithms such as Monte Carlo Tree Search (MCTS) have shown outstanding performance in general video game-playing~\citep{Levinegvgp, perez2016general, sironi2018self}. The ability to play different game variants is important because real-world games are frequently updated by their developers. Therefore, in this work, we focus on search-based methods for strategy game playing.

One of the most challenging problems for search-based algorithms is the combinatorial search space. Unfortunately, strategy games typically have a combinatorial search space. In strategy games such as Starcraft, a number of units (e.g. buildings, and armies) are distributed on the map. The state space of these games is defined as the combination of unit properties (e.g. positions, health points). This combinatorial space increases exponentially with the number of game components (including the unit number and unit property etc.)~\citep{ouessai2020improving, ouessai2020parametric}. On top of that, most strategy video games have a large set of unit variants and each unit has a diverse set of properties. Together, they produce large state and action spaces, resulting in a much larger branching factor compared to other games. With a large branching factor, MCTS finds it difficult to explore the tree deeply for accurate action-value approximation and thus fails to perform well in these games.

State abstraction~\citep{jiang2014improving, hostetler2017sample} is a powerful technique that helps MCTS solve large-scale planning problems. State abstraction methods focus on simplifying the search space, which is often achieved by aggregating similar states. In strategy games, state abstraction~\citep{chung2005monte, synnaeve2012bayesian, dockhorn2021game, xu2023} has been applied to reduce the search space and gain significant performance improvements. However, one of the issues that hinder the application of state abstraction is a lack of data for approximating the state abstraction, resulting in a possible poor-quality state abstraction. To avoid this state abstraction to degrade the performance, Xu et al.~\citep{xu2023} proposed an \textit{early stop} mechanism to abandon the constructed state abstraction at an early stage. However, this approach introduces a hyperparameter whose range depends on the training budget, making it difficult to select an appropriate value.

In this paper, we propose the size-constrained state abstraction (SCSA), a novel approach to address the negative effect of a potential poor-quality state abstraction. SCSA limits the maximal number of nodes in the same node group and does not need the \textit{early stop}. Meanwhile, its hyperparameter is less sensitive to the previous approach. Finally, we evaluate the SCSA agent in $3$ strategy games using a common value of this size limit. It outperforms all the baseline agents in $2$ simple games and achieves results competitive to Elastic MCTS~\citep{xu2023} in another more complex game.

The main contributions of this work are listed below:
\begin{enumerate}
    \item We proposed a novel approach to address planning with a poor-quality state abstraction in strategy game-playing.
    \item Our empirical results show that the proposed method achieves outstanding performance in 3 strategy games of different complexity.
    \item We analyzed the compression rate under the SCSA and Elastic MCTS~\citep{xu2023}. SCSA shows a lower compression rate, revealing a trade-off between memory usage and agent performance under the state abstraction.
\end{enumerate}

\begin{figure} [t]
    \centering
    \includegraphics[width=1.0\columnwidth,trim=0 0cm 0 0cm, clip]{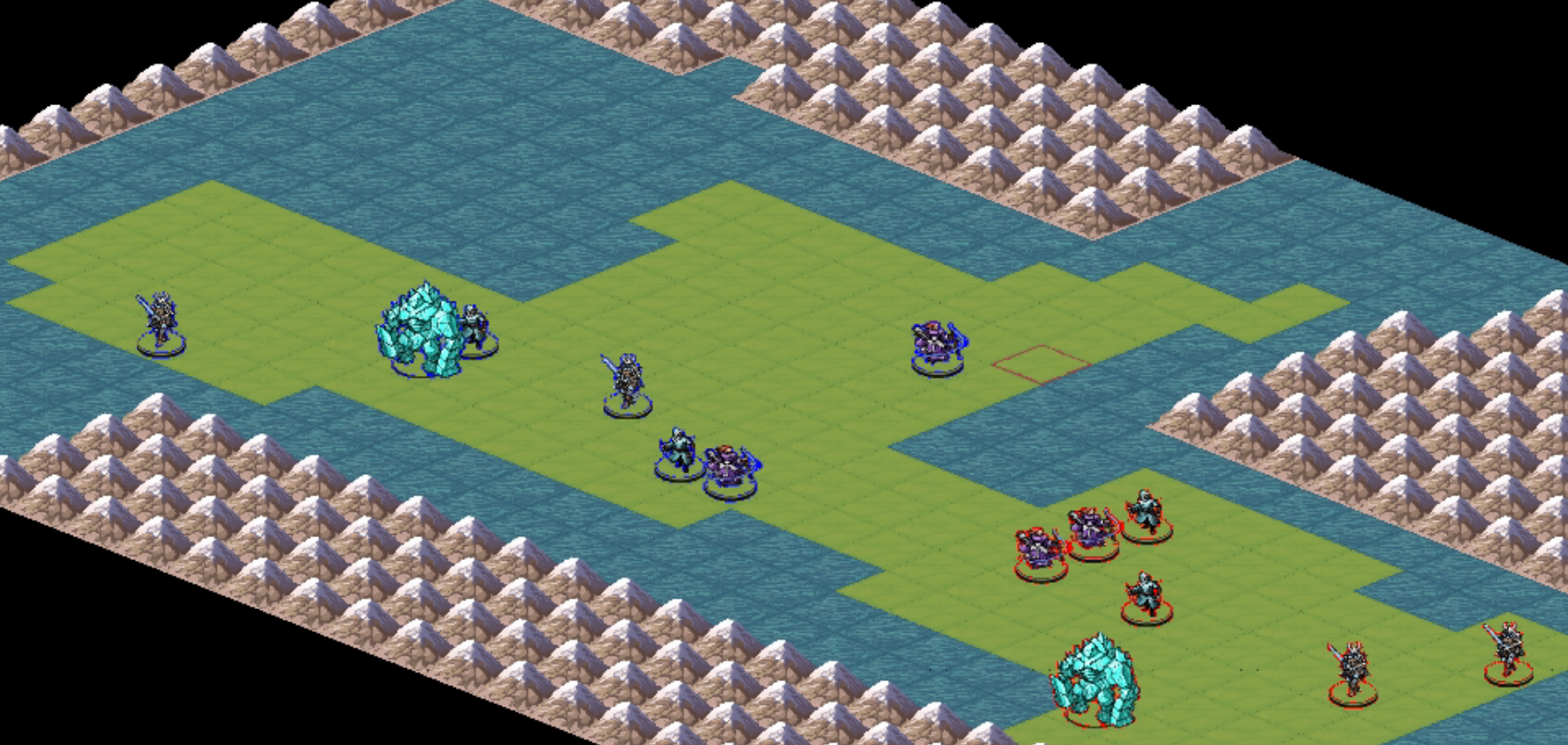}
    \caption{A screenshot of \textit{Kill The King} game. In this case, each player has one \textit{king}, two \textit{warriors}, two \textit{archers} and two \textit{healers}}
    \label{fig:ktk_demo}
\end{figure}
\begin{figure} [t]
    \centering
    \includegraphics[width=1.0\columnwidth,trim=0 0cm 0 0cm, clip]{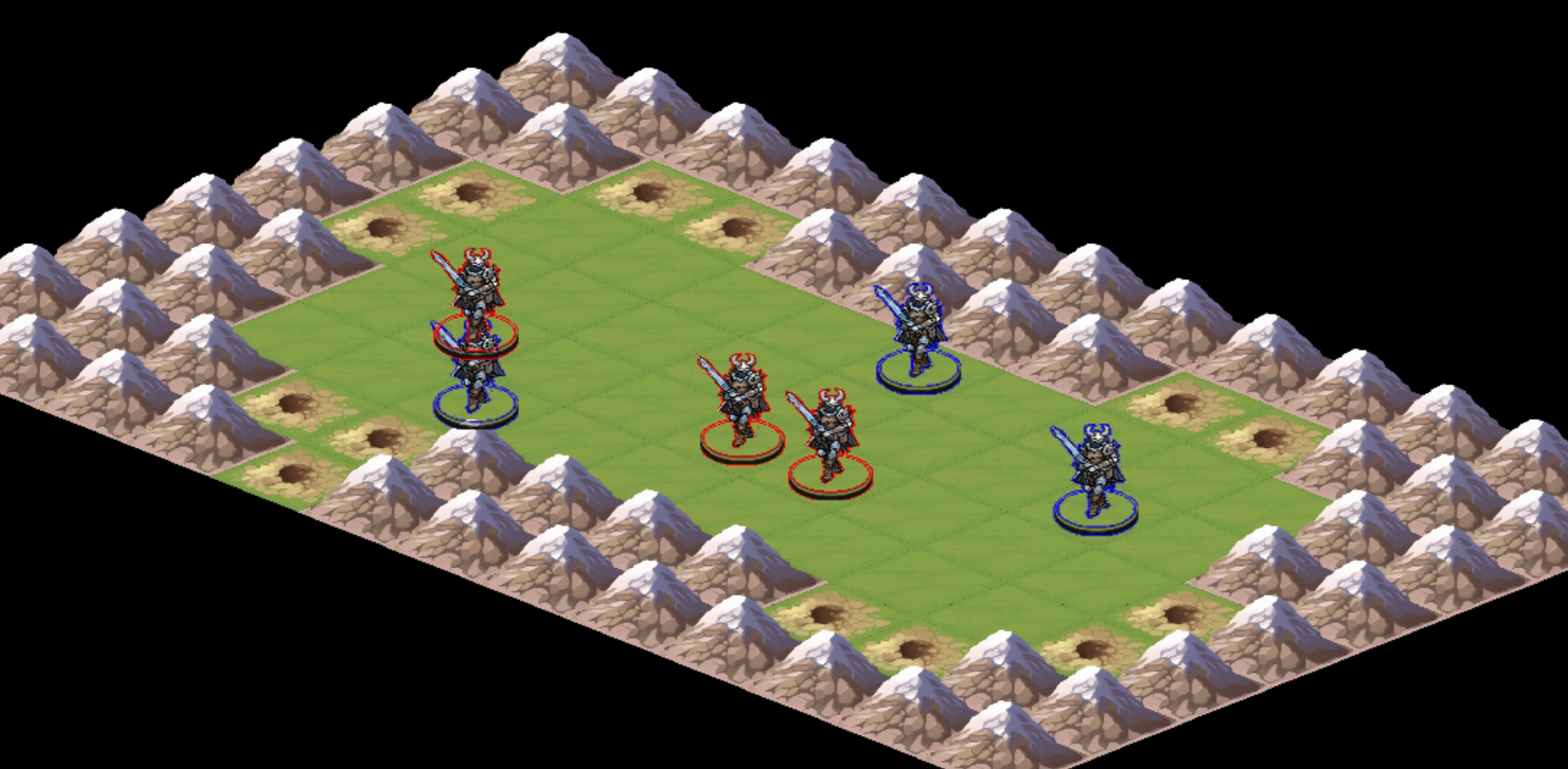}
    \caption{A screenshot of \textit{Push Them All} game. Each player has three \textit{pushers} that try to push enemy units into holes.}
    \label{fig:pta_demo}
\end{figure}
\begin{figure}[t]
    \centering
    \includegraphics[width=1.0\columnwidth,trim=0 0cm 0 0cm, clip]{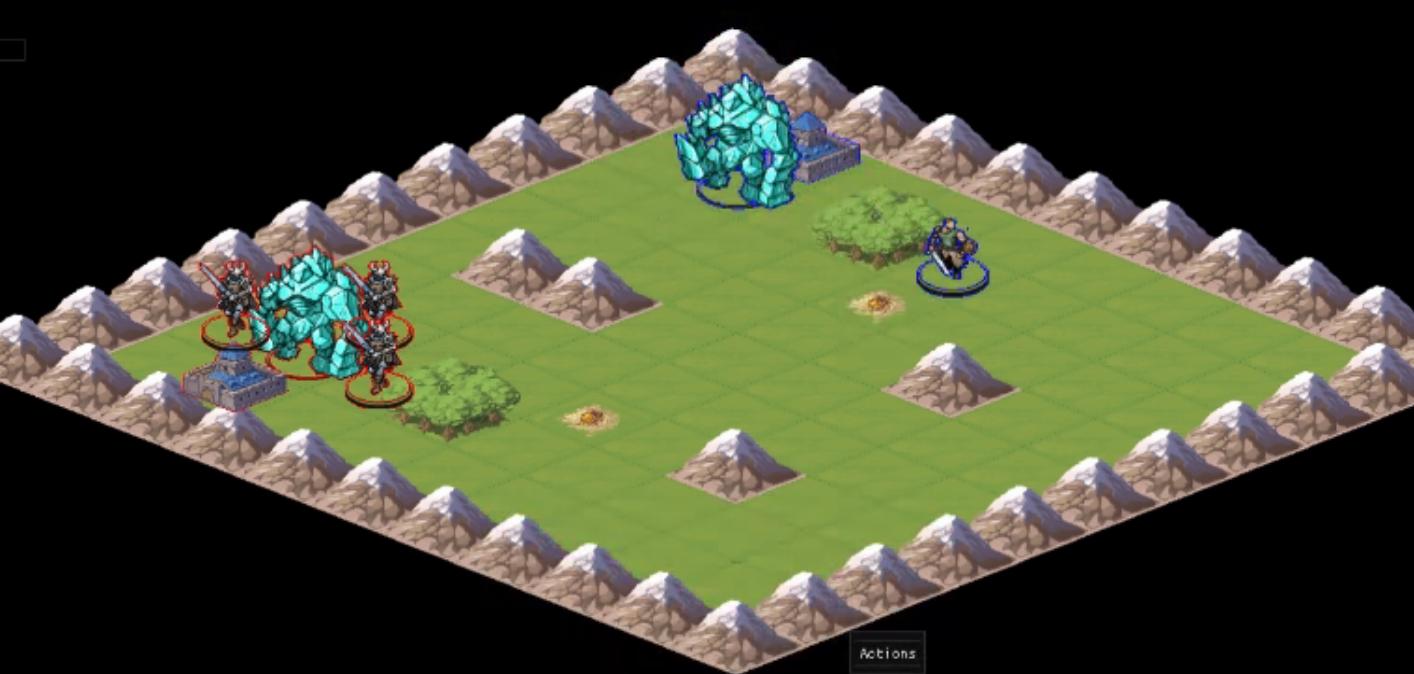}
    \caption{A screenshot of \textit{Two Kingdom} game. The blue side spawned a \textit{worker} to mine the gold and the red side spawned three warriors to protect the \textit{king}.}
    \label{fig:tk_demo}
\end{figure}
\section{Related Work}
State abstraction for MCTS recently gained much interest from the community. Jiang et al.~\citep{jiang2014improving} proposed to aggregate same-layer tree nodes with Markov decision process homomorphism approximated from samples. This method shows a promising performance in the board game Othello. Anand et al.~\citep{anand2015asap} proposed a state-action abstraction method that aggregates state-action pairs instead of states (tree nodes). Anand et al.~\citep{anand2016oga} propose progressive state abstraction that updates the state abstraction more frequently instead of per batch. Hostetler et al.~\citep{hostetler2017sample} proposed a progressive refinement method to construct state abstraction. Baier et al.~\citep{baier2020guiding} proposed \textit{abstraction over opponent moves} to aggregate tree nodes having the same opponent moving history. Sokota et al.~\citep{sokota2021monte} proposed \textit{abstraction refinement} to reject similar states to be added in the tree. These methods prove the effectiveness of state abstraction in tackling large branching factors in MCTS. However, their application is mainly limited to planning problems and board games. This work instead focuses on more complex strategy games.

In the early study, hand-crafted state abstraction were applied to help strategy game play. Chung et al.~\citep{chung2005monte} used a handcrafted state abstraction to divide the game map into tiles. Synnaeve et al.\citep{synnaeve2012bayesian} proposed a mechanism to separate the map in StarCraft to regions that are connected through checkpoints. Uriarte et al.\citep{uriarte2014game} also used the technique developed by Synnaeve et al.~\citep{synnaeve2012bayesian} but further removed combat-irrelevant units from the map. Although these hand-crafted state abstractions can significantly reduce the size of state space for some games, they rely on human heuristics and thus fails to generalize to different games. 

Except for hand-crafted state abstraction, automatic state abstraction are also explored in strategy game-playing in recent years. A parameter optimization method~\citep{lucas2018n} was leveraged to search unit features that can be removed from the unit vectors. By removing some features, states having all other features the same are merged. Dockhorn et al.~\citep{dockhorn2021game} proposed to represent game states with a combination of unit vectors. Our work is closely related to Xu et al.~\citep{xu2023}, where an elastic MCTS method is proposed for strategy game-playing. In elastic MCTS, the state abstraction is first constructed in a batch manner, similar to Jiang et al.~\citep{jiang2014improving}. Later, the constructed state abstraction is abandoned and abstract nodes are split into ground tree nodes. Our work does not need to abandon the state abstraction, and thus is more consistent with state abstraction usage in planning~\citep{jiang2014improving, anand2015asap, anand2016oga, hostetler2017sample}.

\section{The Stratega platform}
Stratega~\citep{dockhorn2020stratega} is a general strategy game platform for testing AI agents. To evaluate the general performance of the proposed method, we select $3$ two-player turn-based strategy games from the Stratega platform. They are \textit{Kill The King (KTK), Push Them All (PTA)} and \textit{Two Kingdoms (TK)}. We next introduce the details of these games.

In \textit{KTK} (Figure~\ref{fig:ktk_demo}), each player controls a set of units including a \textit{king}. The goal of this game is to kill the opponent's \textit{king}. We instantiate the army for each player as a \textit{king}, a \textit{warrior}, an \textit{archer}, and a \textit{healer}. All units have the move action. Based on that, the \textit{king} and the \textit{warrior} can attack neighbour enemy units. The \textit{archer} can attack enemy units in range. The \textit{healer} can heal ally units. Following Xu et al.~\citep{xu2023}, each unit also has an \textit{Do-nothing} action. The action space size for a 4-unit army is about $10^5$.

In \textit{PTA} (Figure~\ref{fig:pta_demo}), a player controls units to push enemy units in different directions. The unit being pushed will move its position toward the corresponding direction. To win this game, all the enemy units need to be pushed into holes distributed in the map. Each player has $3$ \textit{pusher} units. The action set for each \textit{pusher} is \textit{[Move, Push, Do-nothing]}, resulting in an action space of $(4+1) \times 4 \times 4 = 80$. The first term indicates moving in 4 directions or not moving, the second term is selecting a neighbour unit (there are $4$ neighbour grids) and the last term is pushing the enemy unit in $4$ different directions. With $3$ \textit{pushers}, the final action space is $80^3 = 512,000$.

The \textit{TK} game (Figure~\ref{fig:tk_demo}) is more complex. It consists of technologies, resources, unit spawning, and combat. At the beginning of a gameplay, each player has a \textit{castle} and a \textit{king}. The aim is the same as \textit{KTK}, i.e. kill the opponent \textit{king}. However, a set of units need to be spawned from the \textit{castle}. A technology \textit{Mining} is required to be researched for spawning \textit{worker}. The research takes one round to be finished. The \textit{Worker} unit can collect gold from gold veins and \textit{Warrior, Night, Wizard} and \textit{Healer} can be spawned with gold.

\section{Background}

\subsection{Monte Carlo Tree Search}
MCTS~\cite{browne2012survey} is a method to solve sequential decision-making problems with a forward model. The forward model is used to roll out the game. I.e., given a state and a valid action under this state, the forward model returns the next state. Using the forward model, MCTS builds up a tree to approximate the value for actions under the current state. In the generated tree, each node represents a game state and each branch represents a valid action of its source node. We next introduce the 4 stages for building up this tree: \textit{selection, expansion, rollout} and \textit{back-propagation}.

The \textit{selection stage} selects a tree node as an input for the subsequent stages. The \textit{selection} starts from the root node and keeps selecting a branch to the next layer until a target node is reached. The target node could be a leaf node (a node with no children) or a node where not all its actions have been added to the tree as branches. To select the next-layer node, node values (e.g. the UCB value~\citep{kocsis2006bandit}) for all its children are calculated and the node with the highest UCB is selected. Depending on the node type of the target node, MCTS enters different stages. If the target node is a terminal state, it enters the \textit{back-propagation} directly. In another case, the target node is a non-terminal state, an action that has not yet been added as a branch. By running this action in the forward model, the next state is returned and is added as a new child. Based on this new state, a roll-out policy takes a sequence of actions until a pre-determined depth or a terminal state is reached. This is the \textit{rollout} stage. The output state from rollout is evaluated by a state evaluation function to obtain a score. This score is used by the \textit{back-propagation} stage. In the \textit{back-propagation} stage, the score from the target state is added to all states in the trajectory of selection. i.e. a node sequence from the root node to the target node.

Each MCTS iteration consists of these 4 or 3 stages (the roll-out stage is skipped if the selected node is a terminal state). The computation budget in this work is set as the maximum number of forward model calls. After running out of the budget, a recommendation policy selects an action to execute in the game. A common recommendation policy is selecting the branch leading to a node with the highest visit count.

\subsection{Monte Carlo Tree Search with Unit Ordering}
In strategy games where many units are distributed on the map, the action space is the combination of all unit actions, which can easily reach a high complexity. e.g. in \textit{KTK}, the combinatorial action space reaches a magnitude of $10^5$.
To reduce the action space, Xu et al.~\citep{xu2023} propose the MCTS with unit ordering (\MCTSu). In \MCTSu, the move ordering of units is randomly initialized and is fixed throughout the whole game. Each node controls only one unit and its children control the subsequent unit in the move order. With this setting, the tree becomes deeper but narrower.~\MCTSu~has shown a strong performance in the multi-unit strategy games.

\subsection{State Abstraction and Approximate MDP Homomorphism}
A Markov Decision Process (MDP) is defined as $\langle\mathbb{S}, \mathbb{A}, R, P, \gamma \rangle$, where the $\mathbb{S}$ is the state space, $\mathbb{A}$ the action space, $R:\mathbb{S} \times \mathbb{A} \mapsto \mathbb{R}$ the reward function, $P: \mathbb{S} \times \mathbb{A} \mapsto \mathbb{S}$ the transition function and $\gamma \in \mathbb{R}$ is a discount factor. A state abstraction for an MDP is $<\mathbb{S}_\phi, \mathbb{A}, R, \hat{P}, \gamma>$, where the $\mathbb{S}_\phi$ is the abstract state space. Each abstract state includes a set of states. The $\hat{P}: \mathbb{S}_\phi \times \mathbb{A} \mapsto \mathbb{S}_\phi$ defines a transition function based on abstract states.

A key step to construct state abstraction is defining a state mapping function $\phi: \mathbb{S} \mapsto \mathbb{S}_\phi$ that maps a ground state to an abstract state. The function $\phi$ can be implemented by defining similarity between states and aggregating similar states to the same abstract state. Approximate MDP homomorphism~\citep{ravindran2004approximate} is a typical state similarity measurement. For two states $s_1$ and $s_2$, it is defined by the approximate error of reward function $\epsilon_R$ and the approximate error of transition function $\epsilon_T$:
\begin{align}\label{approxerrors}
    &\epsilon_R(s_1,s_2) = \max_{a\in \mathbb{A}} \left|R(s_1, a) - R(s_2, a)\right| \\
    &\epsilon_T(s_1,s_2) = \max_{a\in\mathbb{A}} \sum_{s'_{\phi} \in \mathbb{S}_\phi} \left| \sum_{s' \in s'_{\phi}} T(s'|s_1, a) - \sum_{s' \in s'_{\phi}} T(s'|s_2, a)\right|
\end{align}
where $T(s'|s, a)$ is the transition probability, $\epsilon_R$ measures the maximal difference between reward functions of the given states and $\epsilon_T$ measures the worst-case total variation distance between state transition distributions.

\subsection{Elastic Monte Carlo Tree Search}
The elastic MCTS method~\citep{xu2023} is built upon \MCTSu. It aggregates tree nodes with approximate MDP homomorphism. The constructed node groups are split into ground tree nodes with \textit{early stop}. Algorithm~\ref{alg1} and Algorithm~\ref{alg2} (without the blue parts) provide pseudocode for elastic MCTS. 

For every $B$ MCTS iteration (line $6$ in Algorithm~\ref{alg1}), elastic MCTS checks all the tree nodes that have not yet been added in an abstraction node and calculates their approximate MDP homomorphism errors (line $8$-$9$ in Algorithm~\ref{alg2}). If the errors between the candidate state $s_1$ and an abstraction node $\hat{s}$ are below the pre-determined error thresholds $\eta_R$ and $\eta_T$, $s_1$ is added into $\hat{s}$ (line $13$). If there is no abstract node that matches this condition, a new abstract node is created with the $s_1$ as the only member node (line $15$). The \textit{early stop} shows in line $4$-$5$ from Algorithm~\ref{alg1}. It splits all abstract nodes into ground nodes once the MCTS iteration reaches an \textit{early stop} threshold $\alpha_{ES}$.

\section{Method}
Based on MCTS, our method automatically groups tree nodes by the approximate MDP homomorphism. Following Jiang et al.~\citep{jiang2014improving} and Xu et al.~\citep{xu2023}, SCSA groups tree nodes from the same layer at every batch (a fixed number of MCTS iterations). At each iteration, the MCTS samples one trajectory that consists of a sequence of nodes, starting from the root node to a leaf node. To approximate the MDP homomorphism, a batch of samples is required for calculating the approximate errors (Equation~\ref{approxerrors}). Therefore, for every $B$ iteration(s), SCSA checks every tree node that has not yet joined a node group to expand the current abstraction. There are two approaches for a node to be added to the existing abstraction, depending on the approximate MDP homomorphism errors. If the approximate errors between this candidate node and a node group are below the thresholds, this candidate node is added to the corresponding node group, becoming a \textit{member node} of this group. When the approximate errors between the candidate node and all same-layer groups are found higher than the thresholds, a new node group is created and this node becomes the only member node.

It is found that a large number of samples are required to obtain high-quality state abstractions~\citep{jiang2014improving}. In strategy games where the search space are large, it is infeasible to obtain enough samples. Under limited samples, the constructed state abstraction might be of unstable quality. Moreover, it is difficult to evaluate the quality of the constructed state abstraction. Xu et al.\citep{xu2023} discovered that abandoning the existing state abstraction in the middle of MCTS running can bring significant performance improvement. In contrast to their approach~\citep{xu2023}, SCSA does not abandon the state abstraction. Instead, a global size constraint is defined to limit the maximum number of member nodes for every node group. Below, we introduce the abstraction construction in detail.

The pseudocodes of the SCSA algorithm are shown in Algorithm~\ref{alg1} and Algorithm~\ref{alg2}, with highlighted lines in red representing the removed part from Elastic MCTS, and the lines highlighted in blue are newly introduced by the SCSA method. The computation budget constant is $N_{FM}$, meaning the maximum number of available forward model calls. In Algorithm~\ref{alg1}, lines $6$-$7$ presents the early stop with a threshold $\alpha_{ES}$~\citep{xu2023}. Our method removes this part.

We first introduce the hyperparameters, $N_{FM}$ the computation budget, $B$ the batch size, $\eta_R$ reward function error, $\eta_T$ transition error and \textit{SIZE\_LIMIT} the maximum node group size. In the beginning, the abstraction $\phi$ is initialized by mapping states to themselves. Within the computation budget (line $4$), an MCTS iteration is run with the forward model cost $c_{FM}$ and the current tree depth $L$ returned (line $5$). For every $B$ iteration, the state abstraction is updated by calling the \textit{ConstructAbstraction} function (Algorithm~\ref{alg2}), after which the forward model call counter $n_{FM}$ and the MCTS iteration counter $n_{MCTS}$ are updated.

We next introduce the \textit{ConstructAbstraction} function. Algorithm~\ref{alg2} iterates from the bottom of the tree to the root node, layer by layer (line $2$). For each layer, all nodes that are not added to the abstraction are iterated (line $3$). For a candidate node $s_1$, the algorithm iterates through all same-layer abstract nodes to consider accepting $s_1$ (line $5$). Specifically, SCSA limits the maximal abstract node size. Therefore, if an abstract node is found exceeds the limit, this abstract node is skipped (line $6$-$7$). Otherwise, the approximate errors between $s_1$ and each state from the abstract node is calculated (line $10$-$11$). A node is added into an abstract node (line $14$-$15$) only if the similarities between $s_1$ and all ground nodes from the abstract node are below the thresholds (line $9$-$13$). If a node is finally find not added into any abstract node, a new abstract node is created (line $17$-$18$).

\begin{algorithm}[!t]
\caption{Elastic MCTS
}
\label{alg1}
    \begin{algorithmic}[1]
    \STATE \textbf{Require:} $N_{FM}, B, \eta_{R}, \eta_{T}, K, \textcolor{blue}{\textit{SIZE\_LIMIT}}$ 
    \STATE \textbf{Initialize:} $n_{FM} = 0, n_{MCTS}=0$
    \STATE $\phi := s \rightarrow \hat{s}, \hat{s} = \{s\}$ \ \ \ \ \ \ \ \ \ \ \ \# Initialize the abstraction
    \WHILE{$n_{FM} < N_{FM}$}
    
        \STATE $c_{FM}, L$ = $MCTSIteration(\phi)$
        
        \IF { \color{red}$n_{MCTS} > \alpha_{\textit{ES}}$\color{black}} 
            \color{red} \STATE $\ \ \ \ \phi := s \rightarrow \hat{s}, \hat{s} = \{s\}$ \color{black}
        \ELSIF{ $n_{MCTS} \% B == 0$ }
            \STATE $\phi = ConstructAbstraction(\phi, \eta_{R}, \eta_{T}, L, \textcolor{blue}{\textit{SIZE\_LIMIT})}$
        \ENDIF
        \STATE $n_{FM} = n_{FM} + c_{FM}$
        \STATE $n_{MCTS} = n_{MCTS}+1$
    \ENDWHILE
    \end{algorithmic}
\end{algorithm}

\begin{algorithm}[!t]
\caption{ConstructAbstraction}
\label{alg2}
\begin{algorithmic}[1]
    \STATE \textbf{Require:} $\phi, \eta_{R}, \eta_{T}, L, \textcolor{blue}{\textit{SIZE\_LIMIT}}$
    \FOR{$l=L$ to $1$}
        \FORALL{node $s_1$ in depth $l$ that is not grouped}
            \STATE $s_1\_in\_\phi = \text{false}$
            \FORALL{abstract node $\hat{s}$ in $\phi$}
                \begin{algoblue}
                \IF {$|\hat{s}| > \textit{SIZE\_LIMIT}$ }
                \STATE \textbf{break}
                \ENDIF
                \end{algoblue}
                \STATE $s_1$\_in\_$\hat{s}$ $=$ \text{true}
                \FORALL{node $s_2$ in $\hat{s}$}
                \STATE $\epsilon_R = \max_a |R(s_1, a) - R(s_2, a)| $
                \STATE $\epsilon_T = \sum_{s'} |T(s'|s_1, a) - T(s'|s_2, a)|$
                \IF { $\epsilon_R > \eta_R $ \textbf{or} $ \epsilon_T > \eta_T $}
                \STATE $s_1$\_in\_$\hat{s}$ $=$ false, \textbf{break}
                \ENDIF
                \ENDFOR
            \IF { $s_1$\_in\_$\hat{s}$ $==$ true}
            \STATE Add  $s_1$ in abstract node $\hat{s}$
            \STATE $s_1\_in\_\phi = true$
            \ENDIF
            \ENDFOR
        \IF {$s_1\_in\_\phi == \text{false}$}
        \STATE $\phi(s_1) = \{ s_1\}$\ \ \ \ \# Create a new abstract node
        \ENDIF 
        \ENDFOR
    \ENDFOR
\end{algorithmic}
\end{algorithm}

\section{Experiments}
\textbf{Baselines:} We implement 5 baseline agents to evaluate the performance of the SCSA agent. They are \textit{\rb, MCTS, \MCTSu, \rMCTSu} and \textit{\eMCTSu}. Details about each agent are listed below:
\begin{enumerate}
    \item \textit{\rb}: Stratega platform has implemented a \rb agent for each game. We here briefly introduce their implementation. The \rb agent for \textit{KTK} prioritizes attacking isolated enemy units and healing strong ally units. For each enemy unit, an isolation score is calculated considering its nearby ally units and enemy units. At each round, the \rb agent controls its units to i) approach the enemy units with the highest isolation score and attack them; and ii) approach an ally unit to heal it. The \textit{PTA} \rb agent controls \textit{pushers} to approach the nearest enemy unit and push it towards the nearest hole. In \textit{TK}, the \rb agent first researches \textit{Mining}, which is necessary for spawning \textit{workers}. Once the research is finished, a \textit{worker} is spawned and is assigned the task of collecting gold from the nearest gold vein. These gold are used to spawn \textit{warriors}. Once the number of \textit{warriors} reaches $2$, these \textit{warriors} are sent to attack the enemy \textit{king}. Whenever its \textit{warriors} died, new \textit{warriors} would be spawned if they had enough gold resources.
    \item \textit{MCTS}: An MCTS agent without using state abstraction.
    \item \textit{\MCTSu}: An MCTS agent with unit ordering.
    \item \textit{\rMCTSu}: An MCTS$_u$ agent with randomized state abstraction. Each new tree node either joins an existing node group (with the probability $\frac{1}{N+1}$ for each group, supposing there are already $N$ node groups) or creates a new node group with itself as the only \textit{member node} (with the probability $\frac{1}{N+1}$).
    \item \textit{\eMCTSu}: MCTS$_u$ with state abstraction based on approximate MDP homomorphism and \textit{early stop}~\citep{xu2023}.
    \item \textit{SCSA}: An MCTS$_u$ agent with approximate MDP homomorphism abstraction. Each abstract node has a size limit defined by \textit{SIZE\_LIMIT}.
\end{enumerate}

\textbf{Heuristic functions:} The same as typical MCTS in games, we utilize heuristic functions to evaluate states reached by the MCTS roll-out. The heuristic functions are game-specific. In each game, all agents except for the \rb agent share the same heuristic function. Below, we introduce the implementation details of the heuristic functions for each game.

In all games, the scores of states where the player wins, loses and draws the game are $1$, $-1$ and $0$, respectively. For all the other states, the heuristic function returns a score between $0$ and $1$. The \textit{KTK} heuristic function returns a score of $R = 1-\frac{d \cdot h}{D_{\text{ktk}} \cdot H}$, where the $d$ is the sum of the distance from each ally unit to the enemy king, $D_{\text{ktk}}$ is the maximum value of $d$, $h$ is the health points of the enemy king and $H$ is the maximum value of $h$. The strategy is controlling the units to approach the enemy king and try to search for a state that leads to victory.

For \textit{PTA}, the score of a state is a sum of three parts. The first one is $0.2 \times \frac{\sum_u \min_{u'} dis(u, u')}{D_{\text{pta}}}$, where the $u$ is an ally unit, $u'$ is an enemy unit, $dis(\cdot, \cdot)$ returns the Euclidean distance between the two units. The second part is $0.4 \times \frac{|U_t|}{|U_0|}$, where $|U_t|$ is the number of alive ally units at time step $t$ and $U_0$ is the number of the units in the beginning. The last part is $0.4 \times \frac{|U'_0|- |U'_t|}{|U'_0|}$, where $|U'_t|$, $|U'_0|$ are the number of enemy units at time step $t$, respectively.

In \textit{TK}, the state score is calculated according to finishing a series of tasks. Finishing \textit{Mining} research returns $0.2$, having $worker$ alive returns $0.1$ and having units that have action \textit{attack} returns $0.1$. Other scores include $0.1\times$ the distance of ally workers to its nearest gold vein, $0.2\times$ collected gold, $0.3\times$ the distance between all ally units and enemy units. The normalized score lands in $[0,1]$.

\begin{table}[]
\caption{
Agent parameters for Section~\ref{sec:exp1} experiment
}
\centering
\begin{tabular}{c|rrrrrr}
\toprule
\multicolumn{1}{c|}{Agents}       & \multicolumn{1}{l}{C}    & \multicolumn{1}{r}{K}    & \multicolumn{1}{r}{$\alpha_{\textit{ABS}}$} & \multicolumn{1}{r}{$\eta_R$} & \multicolumn{1}{r}{$\eta_T$} & \multicolumn{1}{r}{\textit{SIZE\_LIMIT} } \\ \midrule
\multicolumn{6}{c}{Kill The King (KTK)} \\ \midrule
\multicolumn{1}{r|}{MCTS}         & \multicolumn{1}{r}{0.1} & \multicolumn{1}{r}{10} & \multicolumn{1}{r}{/}               & \multicolumn{1}{r}{/}         & \multicolumn{1}{r}{/}  & \multicolumn{1}{r}{/}        \\ 
\multicolumn{1}{r|}{\MCTSu}         & \multicolumn{1}{r}{1.0} & \multicolumn{1}{r}{10} & \multicolumn{1}{r}{/}               & \multicolumn{1}{r}{/}         & \multicolumn{1}{r}{/} & \multicolumn{1}{r}{/}        \\ 
\multicolumn{1}{r|}{\rMCTSu}         & \multicolumn{1}{r}{0.1} & \multicolumn{1}{r}{10} & \multicolumn{1}{r}{8}               & \multicolumn{1}{r}{/}         & \multicolumn{1}{r}{/}   & \multicolumn{1}{r}{/}      \\
\multicolumn{1}{r|}{\eMCTSu}         & \multicolumn{1}{r}{0.1} & \multicolumn{1}{r}{10} & \multicolumn{1}{r}{10}               & \multicolumn{1}{r}{0.05}         & \multicolumn{1}{r}{1.0}     & \multicolumn{1}{r}{/} \\
\multicolumn{1}{r|}{\SCSA}         & \multicolumn{1}{r}{0.1} & \multicolumn{1}{r}{10} & \multicolumn{1}{r}{/}               & \multicolumn{1}{r}{0.05}         & \multicolumn{1}{r}{1.0}       & \multicolumn{1}{r}{2}  \\
 \midrule

\multicolumn{6}{c}{Push Them All (PTA)}  \\ \midrule
\multicolumn{1}{r|}{MCTS}         & \multicolumn{1}{r}{10} & \multicolumn{1}{r}{10} & \multicolumn{1}{r}{/}               & \multicolumn{1}{r}{/}         & \multicolumn{1}{r}{/}        & \multicolumn{1}{r}{/} \\ 
\multicolumn{1}{r|}{\MCTSu}         & \multicolumn{1}{r}{10} & \multicolumn{1}{r}{20} & \multicolumn{1}{r}{/}               & \multicolumn{1}{r}{/}         & \multicolumn{1}{r}{/}         & \multicolumn{1}{r}{/} \\ 
\multicolumn{1}{r|}{\rMCTSu}         & \multicolumn{1}{r}{0.1} & \multicolumn{1}{r}{10} & \multicolumn{1}{r}{4}               & \multicolumn{1}{r}{/}         & \multicolumn{1}{r}{/}     & \multicolumn{1}{r}{/}    \\
\multicolumn{1}{r|}{\eMCTSu}         & \multicolumn{1}{r}{10} & \multicolumn{1}{r}{10} & \multicolumn{1}{r}{8}               & \multicolumn{1}{r}{1.0}         & \multicolumn{1}{r}{1.0}        & \multicolumn{1}{r}{/} \\
\multicolumn{1}{r|}{\SCSA}         & \multicolumn{1}{r}{10} & \multicolumn{1}{r}{10} & \multicolumn{1}{r}{/}               & \multicolumn{1}{r}{1.0}         & \multicolumn{1}{r}{1.0}   & \multicolumn{1}{r}{2}      \\
\midrule

\multicolumn{6}{c}{Two Kingdoms (TK)}  \\ \midrule
\multicolumn{1}{r|}{MCTS}         & \multicolumn{1}{r}{0.1} & \multicolumn{1}{r}{20} & \multicolumn{1}{r}{/}               & \multicolumn{1}{r}{/}         & \multicolumn{1}{r}{/}       & \multicolumn{1}{r}{/}  \\ 
\multicolumn{1}{r|}{\MCTSu}         & \multicolumn{1}{r}{1.0} & \multicolumn{1}{r}{20} & \multicolumn{1}{r}{/}               & \multicolumn{1}{r}{/}         & \multicolumn{1}{r}{/}    & \multicolumn{1}{r}{/}     \\ 
\multicolumn{1}{r|}{\rMCTSu}         & \multicolumn{1}{r}{0.1} & \multicolumn{1}{r}{10} & \multicolumn{1}{r}{8}               & \multicolumn{1}{r}{/}         & \multicolumn{1}{r}{/}   & \multicolumn{1}{r}{/}      \\ 
\multicolumn{1}{r|}{\eMCTSu}         & \multicolumn{1}{r}{1.0} & \multicolumn{1}{r}{20} & \multicolumn{1}{r}{6}               & \multicolumn{1}{r}{0.05}         & \multicolumn{1}{r}{1.0}  & \multicolumn{1}{r}{/} \\
\multicolumn{1}{r|}{\SCSA}         & \multicolumn{1}{r}{1.0} & \multicolumn{1}{r}{20} & \multicolumn{1}{r}{/}               & \multicolumn{1}{r}{0.05}         & \multicolumn{1}{r}{1.0}  
 & \multicolumn{1}{r}{2} \\ 
 \bottomrule
\end{tabular}
\label{Tab:params}
\end{table}

\subsection{Agent Parameter Optimisation with NTBEA}
As each agent has different optimal parameters for each game, we apply the N-Tuple Bandit Evolutionary Algorithm (NTBEA)~\citep{lucas2018n} to automatically optimize agent parameters in different games. 
The NTBEA uses an N-Tuple system to break down the combinatorial space of the parameters. NTBEA has its own parameters, an exploration factor, the number of neighbours and the number of iterations. Following Xu et al.~\citep{xu2023}, these values are set to $2, 50$ and $50$, respectively. Next, we introduce the parameter space for each game-playing agent.

The parameters for MCTS and \MCTSu~are the exploration factor $C \in \{0.1, 1, 10, 100\}$ and rollout length $K\in \{10, 20, 40\}$. \rMCTSu has $C, K$ and an \textit{early stop} threshold $\alpha_{ES} \in \{4\times B, 8\times B, 10\times B, 12\times B\}$, where the $B=20$ is a constant batch size. \eMCTSu~has $C, K, \alpha_{ES}$ and approximate errors for reward function and transition function $\eta_R \in \{0.0, 0.04, 0.1, 0.3, 0.5, 1.0\}, \eta_T \in \{0.0, 0.5, 1.0, 1.5, 2.0\}$. For SCSA, we use the same parameters as \eMCTSu. SCSA does not require the early stop threshold but requires a \textit{SIZE\_LIMIT}, which is linearly searched and set to $2$ in the first group of experiments. The hyper-parameters tuned by NTBEA are shown in Table~\ref{Tab:params}.

Except for parameters, the budgets for search-based agents vary in different games. We set this budget based on the competitive performance of \MCTSu~playing against the corresponding \rb agent. In \textit{KTK} and \textit{PTA} the budget is set to $10,000$ number of forward model calls. In \textit{TK}, the budget is $5,000$ forward model calls.

\begin{figure}
    \centering
    \includegraphics[width=\columnwidth,trim=0 0cm 0 0cm, clip]{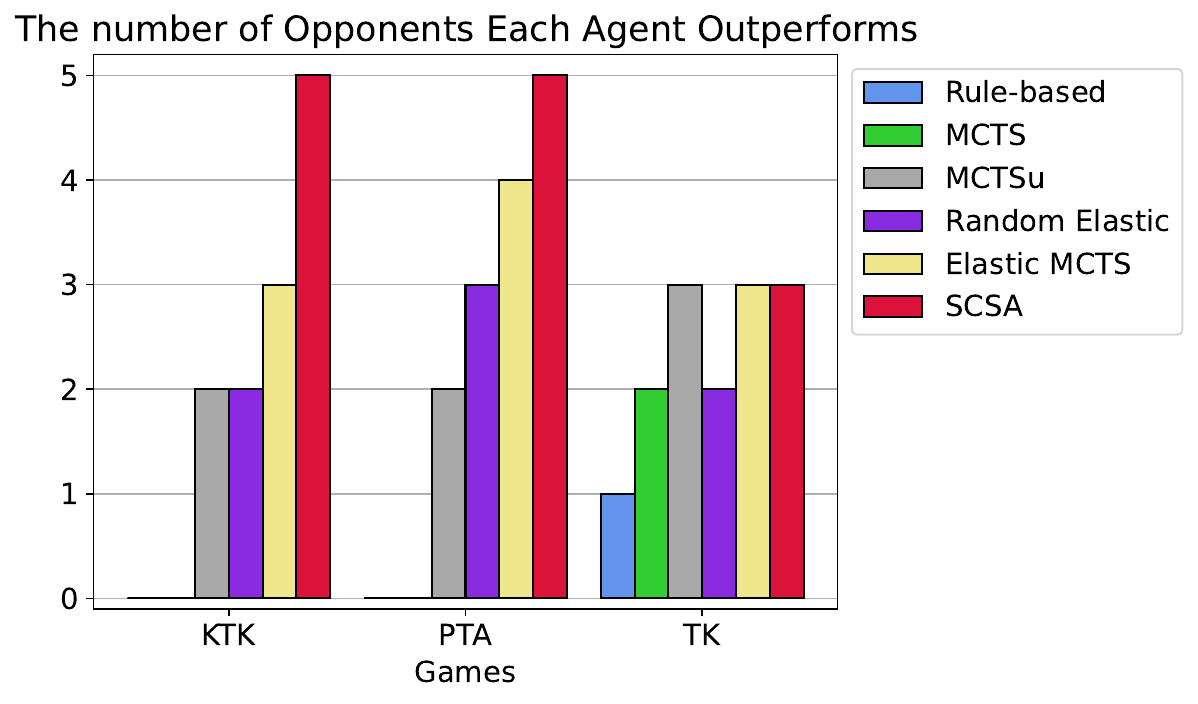}
    \caption{The number of opponents that the agent outperforms.}
    \label{fig:general_performance}
\end{figure}

\subsection{Performance on multi-unit-grid-based games} \label{sec:exp1}
To evaluate the general performance of SCSA agents, we ran an experiment with agents playing against each other in a two-player manner. For each game, we set up 50 initial unit positions (randomly sampled). For each initial unit position, 2 evaluations are made by switching sides. Each evaluation is made with 5 random seeds, resulting in $500$.

The general performance of each agent is shown in Figure~\ref{fig:general_performance}. In KTK and PTA, \SCSA outperforms all its opponents. In the more complex \textit{TK} game, SCSA shows a competitive performance to Elastic~\MCTSu.

The detailed win rates for each agent pair are shown in Tables~\ref{Tab:ktk},~\ref{Tab:pta} and~\ref{Tab:tk}. In \textit{KTK}, the SCSA agent outperforms all its opponent agents in a significant gap. MCTS shows a weaker performance than the rule-based agent. MCTS$_u$~shows a stable and better performance than the MCTS. Elastic~MCTS$_u$ outperforms both \rMCTSu and MCTS$_u$.

In \textit{PTA}, the overall win rates are higher than \textit{KTK}. In this game, the SCSA agent also outperforms all its opponents. Elastic MCTS$_u$ shows a strong performance in that it beats all agents except for the SCSA agent. MCTS$_u$ outperforms MCTS by a large margin. The MCTS agent is still weaker than the rule-based agent.

In \textit{TK}, the \rb agent outperforms the MCTS significantly while other agents outperform \rb with large margins. In this complex game, MCTS$_u$, Elastic MCTS$_u$ and SCSA are showing close performances.

In conclusion, these experiments verify the performance improvement brought by the unit ordering and Elastic MCTS. It also evaluates the performance of the SCSA agent, confirming its outstanding performance in all three games. In addition, it shows good performance of SCSA agents in different games can be achieved by the same value for \textit{SIZE\_LIMIT}. In this experiment, we show that $2$ is an appropriate value for \textit{SIZE\_LIMIT}. Comparing different games, the SCSA agent performs less strongly in the more complex \textit{TK} game, indicating a potential issue of scalability.

\subsection{Influence of abstract state size}
\begin{figure*}
    \subfloat[\normalsize{Kill The King (\textit{KTK})}]{
		\centering
        \includegraphics[width=0.6\columnwidth,trim=0 0cm 0 0.6cm, clip]{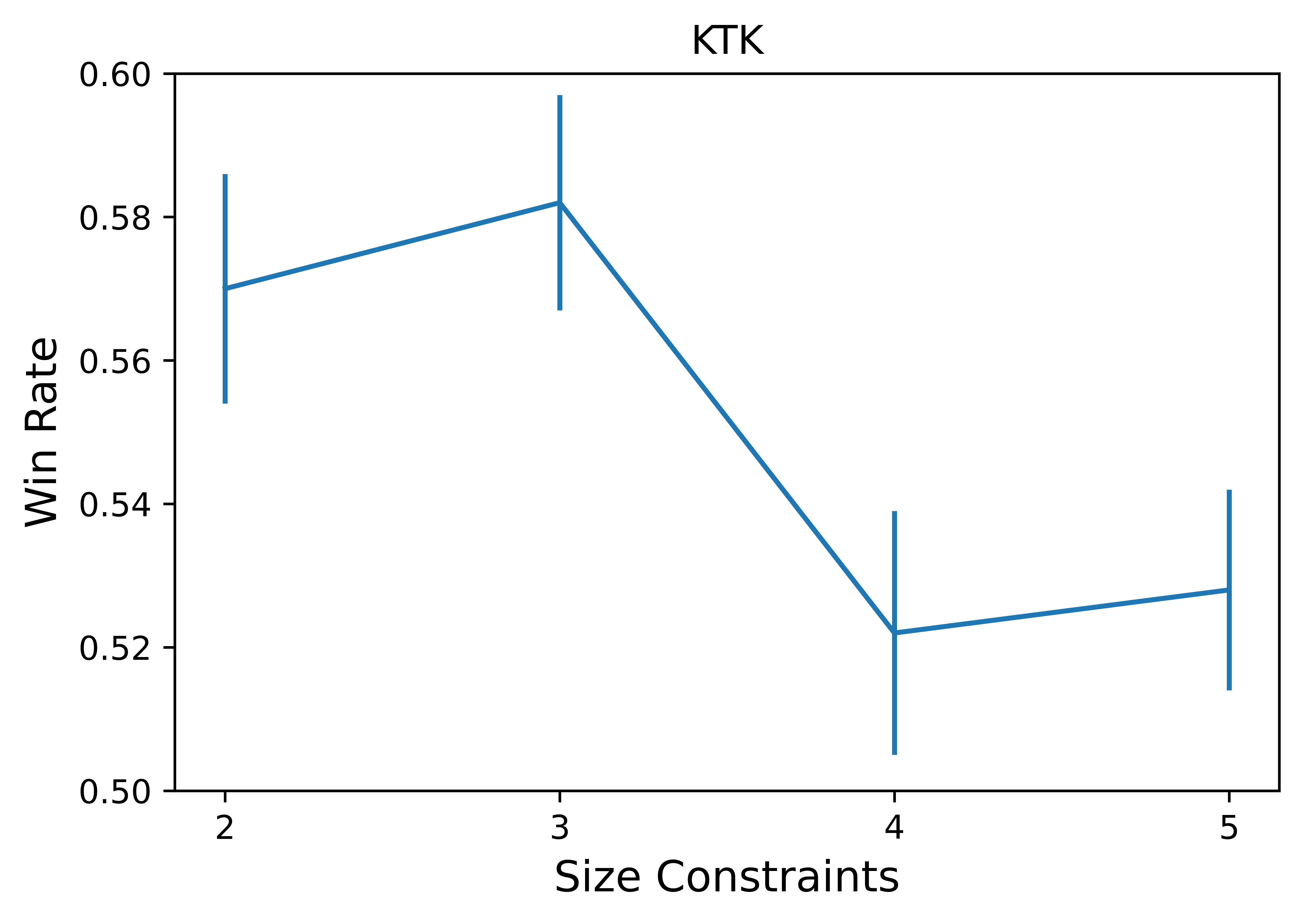}
        \label{fig:ktk_sl}
	}\hfill
    \subfloat[\normalsize{Push Them All (\textit{PTA})}]{
		\centering
        \includegraphics[width=0.6\columnwidth,trim=0 0cm 0cm 0.6cm, clip]{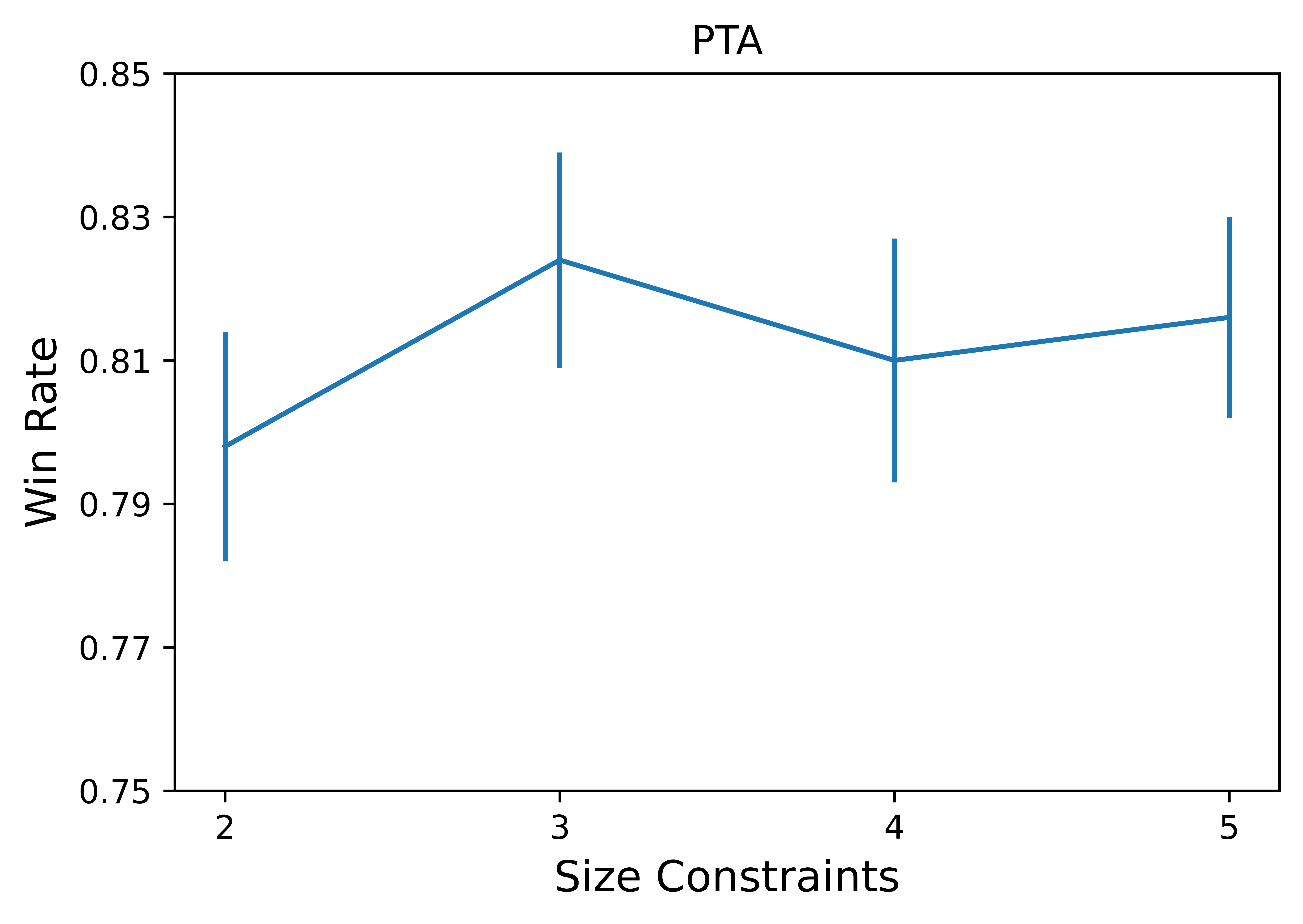}
        \label{fig:pta_sl}
	}\hfill
    \subfloat[\normalsize{Two Kingdoms (\textit{TK})}]{
		\centering
        \includegraphics[width=0.6\columnwidth,trim=0 0cm 0 0.6cm, clip]{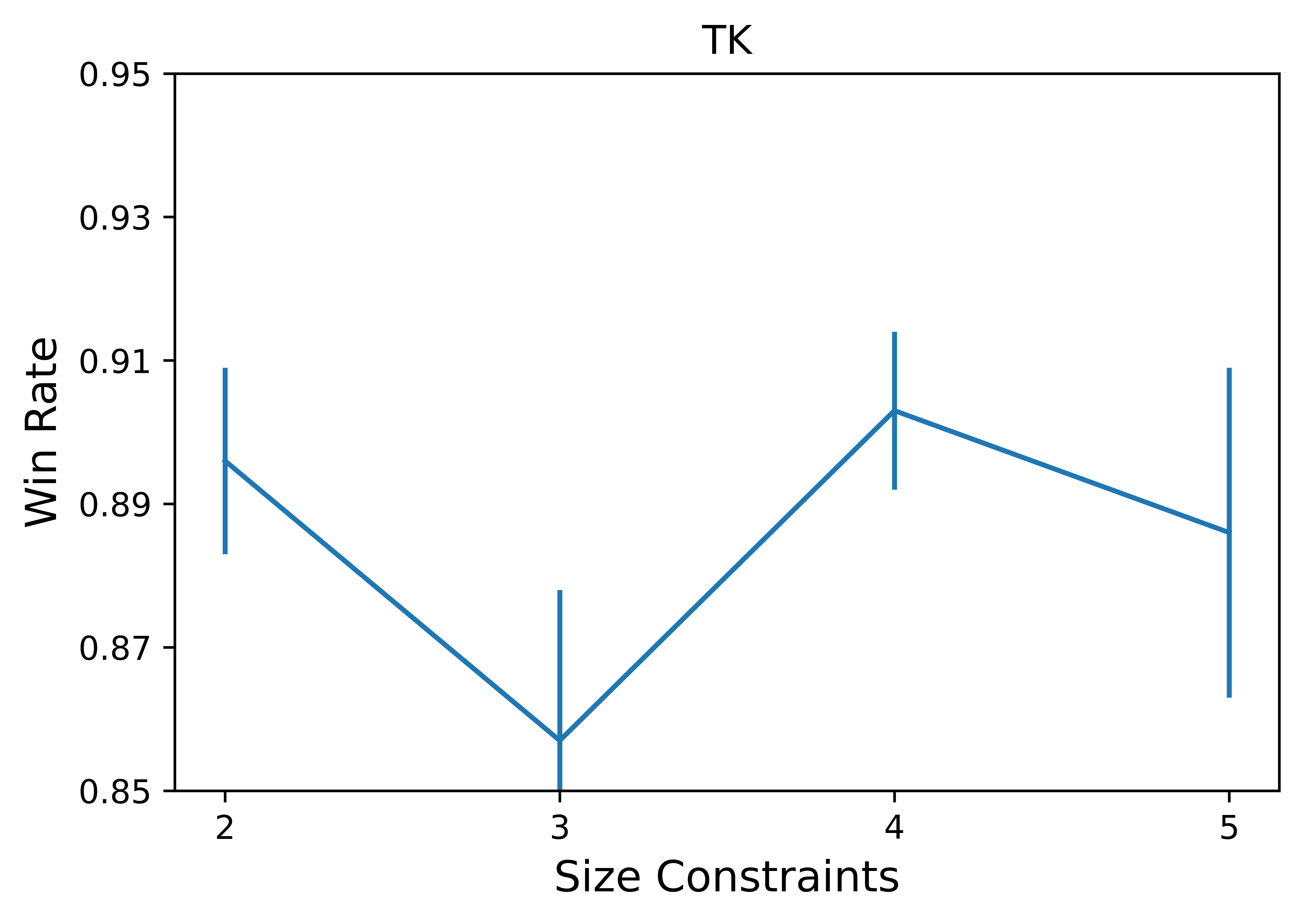}
        \label{fig:tk_sl}
	}
	\caption{Performance of the SCSA agent with different values for its size constraint. Results of the SCSA agent playing against the corresponding \rb agents are visualized, including win rates and standard errors.}
\end{figure*}

To better investigate the influence of different values for \textit{SIZE\_LIMIT}, we assigned different values from $2$ to $5$ and run the agent pair of SCSA - \rb agent. The same as in Section~\ref{sec:exp1}, 500 games are run for each value of \textit{SIZE\_LIMIT}. Figure~\ref{fig:ktk_sl}-\ref{fig:tk_sl} shows the win rates with standard errors in three games. We observe the optimal \textit{SIZE\_LIMIT} values for \textit{KTK} and \textit{PTA} is $3$ but \textit{TK} has its optimal values at $2$ and $4$. We also observed that larger \textit{SIZE\_LIMIT} values (e.g. a value of $5$) can degrade the performance, which reveals the trade-off between the tree size and the performance. With a larger \textit{SIZE\_LIMIT}, more groups are aggregated together therefore the tree size becomes smaller. However, a tree that is too small might cause performance degradation.

We used a value of $2$ for all games and the agents showed satisfactory performance. Compared to different $\alpha_{ES}$ values are required in each domain (See Table V in Xu et al.~\citep{xu2023}), the \textit{SIZE\_LIMIT} is less sensitive across domains.

\begin{table}
\caption{Win rates with standard errors for games \textit{Kill The King} }
\centering
\begin{tabular}{rr|rr}
\toprule
Agent 1 & Agent 2 & Agent 1 & Agent 2 \\ \midrule
\multicolumn{4}{c}{1 King, 1 Archer, 1 Warrior and 1 Healer }\\ \midrule
MCTS& \rb                  &   $47.2 (1.9)\%$    &    $52.8 (1.9)\%$                  \\ 
\MCTSu& \rb    &   $\mathbf{62.2 (1.1)\%}$    &    $37.6 (1.3)\%$                 \\ 
\rMCTSu& \rb   &       $ \mathbf{63.4(1.0)\%} $        &    $36.6 (1.0)\%$                 \\
\eMCTSu& \rb    &       $ \mathbf{54.2(1.4) \%} $        &    $ 44.8(1.6)\%$                 \\ 
\SCSA (ours) & \rb   & $ \mathbf{55.6 (0.6)\%} $ &    $ 44.4(0.6)\%$                 \\ \midrule
\MCTSu & MCTS   &       $ \mathbf{58.0(1.2)\%} $        &    $41.4 (1.2)\%$                 \\
\rMCTSu& MCTS   &       $ \mathbf{61.6 (0.9)\%} $        &    $38.4 (0.9)\%$                 \\
\eMCTSu & MCTS   &       $\mathbf{ 61.4(1.2)\%} $        &    $ 36.6(1.2)\%$                 \\
\SCSA (ours) & MCTS   &       $  \mathbf{58.2(1.9)\%} $        &    $ 30.2 (1.7)\%$      \\ \midrule
\rMCTSu & \MCTSu           &       $49.0 (0.9)\%$     &        $51.0 (0.9)\%$            \\ 
\eMCTSu & \MCTSu   &       $\mathbf{ 61.2(1.4)\%} $        &    $ 38.8(1.4)\%$                 \\ 
\SCSA (ours) & \MCTSu   &       $\mathbf{ 53.8(1.7)\%} $   &    $ 38.8(1.7)\%$                 \\  \midrule
\eMCTSu & \rMCTSu           &       $ 50.2(1.0)\%$     &        $ 49.8(1.0)\%$            \\ 
\SCSA (ours) & \rMCTSu           &       $ \mathbf{52.4(1.4)\%}$     &        $ 47.4(1.3)\%$            \\  \midrule
\SCSA (ours)  & \eMCTSu           &       $ \mathbf{49.2(2.1)\%}$     &        $ 38.8(1.9)\%$            \\ 
 \bottomrule
\end{tabular}
\label{Tab:ktk}
\end{table}

\begin{table}
\caption{Win rates with standard errors for games \textit{Push Them All}}
\centering
\begin{tabular}{rr|cc}
\toprule
Agent 1 & Agent 2 & Agent 1 & Agent 2 \\ \midrule
MCTS& \rb                  &   $48.8 (1.6)\%$    &    $51.2 (1.6)\%$                  \\ 
\MCTSu& \rb    &   $\mathbf{69.0 (1.1)\%}$    &    $30.8 (1.2)\%$                 \\ 
\rMCTSu& \rb   &       $\mathbf{74.0 (2.3)\%} $        &    $26.0(2.3)\%$                 \\
\eMCTSu& \rb    &       $ \mathbf{81.8(0.9) \%} $        &    $ 18.0(1.1)\%$                 \\ 
\SCSA (ours)& \rb    &       $ \mathbf{79.8(1.6) \% }$        &    $ 20.2(1.6)\%$                 \\  \midrule
\MCTSu & MCTS   &       $\mathbf{64.4 (1.1)\%} $        &    $33.6(1.4)\%$                 \\
\rMCTSu& MCTS   &       $\mathbf{86.2 (0.7)\%} $        &    $12.0 (0.7)\%$      \\
\eMCTSu & MCTS   &       $\mathbf{ 85.4(1.5)\%} $        &    $ 13.0(1.7)\%$                 \\
\SCSA (ours) & MCTS   &       $\mathbf{ 86.0(1.4)\%} $        &    $ 12.8(0.9)\%$                 \\ \midrule
\rMCTSu & \MCTSu           &       $\mathbf{73.4 (2.0)\%}$     &        $25.6 (1.9)\%$            \\ 
\eMCTSu & \MCTSu   &       $\mathbf{ 80.2(1.0)\%} $        &    $ 18.0(1.0)\%$                 \\ 
\SCSA (ours) & \MCTSu   &       $\mathbf{ 77.2(1.8)\%} $        &    $ 22.0(2.0)\%$                 \\  \midrule
\eMCTSu & \rMCTSu           &       $ \mathbf{62.4(1.4)\%}$     &        $ 35.8(1.3)\%$            \\ 
\SCSA (ours) & \rMCTSu           &       $ \mathbf{58.8(1.8)\%}$     &        $ 40.4(2.0)\%$            \\   \midrule
\SCSA (ours) & \eMCTSu           &       $ \mathbf{52.0(1.6)}\%$     &        $ 46.8 (1.9)\%$            \\ 
\bottomrule
\end{tabular}
\label{Tab:pta}
\end{table}

\begin{table}
\caption{Win rates with standard errors for games \textit{Two Kingdoms}}
\centering
\begin{tabular}{rr|cc}
\toprule
Agent 1 & Agent 2 & Agent 1 & Agent 2 \\ \midrule
MCTS& \rb                  &   $12.6 (0.5)\%$    &    $\mathbf{81.2 (1.2)\%}$                  \\ 
\MCTSu& \rb    &   $\mathbf{90.8 (1.7)\%}$    &    $7.6 (1.6)\%$                 \\ 
\rMCTSu& \rb   &       $\mathbf{88.0 (1.1)\%} $        &    $11.8(1.2)\%$                 \\
\eMCTSu& \rb    &       $ \mathbf{89.2(1.4) \%} $        &    $ 9.6(1.3)\%$                 \\ 
\SCSA (ours)& \rb    &       $ \mathbf{88.8(1.0) \% }$        &    $ 9.8(1.3)\%$                 \\  \midrule
\MCTSu & MCTS   &       $\mathbf{96.0(0.4)\%} $        &    $4.0(0.4)\%$                 \\
\rMCTSu& MCTS   &       $\mathbf{89.6 (1.2)\%} $        &    $10.4 (1.2)\%$      \\
\eMCTSu & MCTS   &       $\mathbf{ 96.0(0.5)\%} $        &    $ 4.0(0.5)\%$                 \\
\SCSA (ours) & MCTS   &       $\mathbf{ 94.0 (0.6)\%} $        &    $ 6.0(0.6)\%$                 \\ \midrule
\rMCTSu & \MCTSu           &       $43.6 (1.6)\%$     &        $\mathbf{56.2 (1.6)\%}$            \\ 
\eMCTSu & \MCTSu   &       $ 52.2(2.1)\% $        &    $ 47.6(1.9)\%$                 \\ 
\SCSA (ours) & \MCTSu   &       $ 49.0(1.8)\% $        &    $ 50.8(1.8)\%$                 \\  \midrule
\eMCTSu & \rMCTSu           &       $ \mathbf{62.4(1.0)\%}$     &        $ 37.6(1.0)\%$            \\ 
\SCSA (ours) & \rMCTSu           &       $ \mathbf{57.0(1.6)\%}$     &        $ 43.0(1.6)\%$            \\   \midrule
\SCSA (ours) & \eMCTSu           &       $ 51.4(2.7)\%$     &        $ 48.6 (2.7)\%$            \\ 
\bottomrule
\end{tabular}
\label{Tab:tk}
\end{table}

\begin{figure*} [h] \label{fig:cr_compare}
    \subfloat[\normalsize{Kill The King (\textit{KTK})}]{
		\centering
        \includegraphics[width=0.6\columnwidth,trim=0 0cm 0 0.92cm, clip]{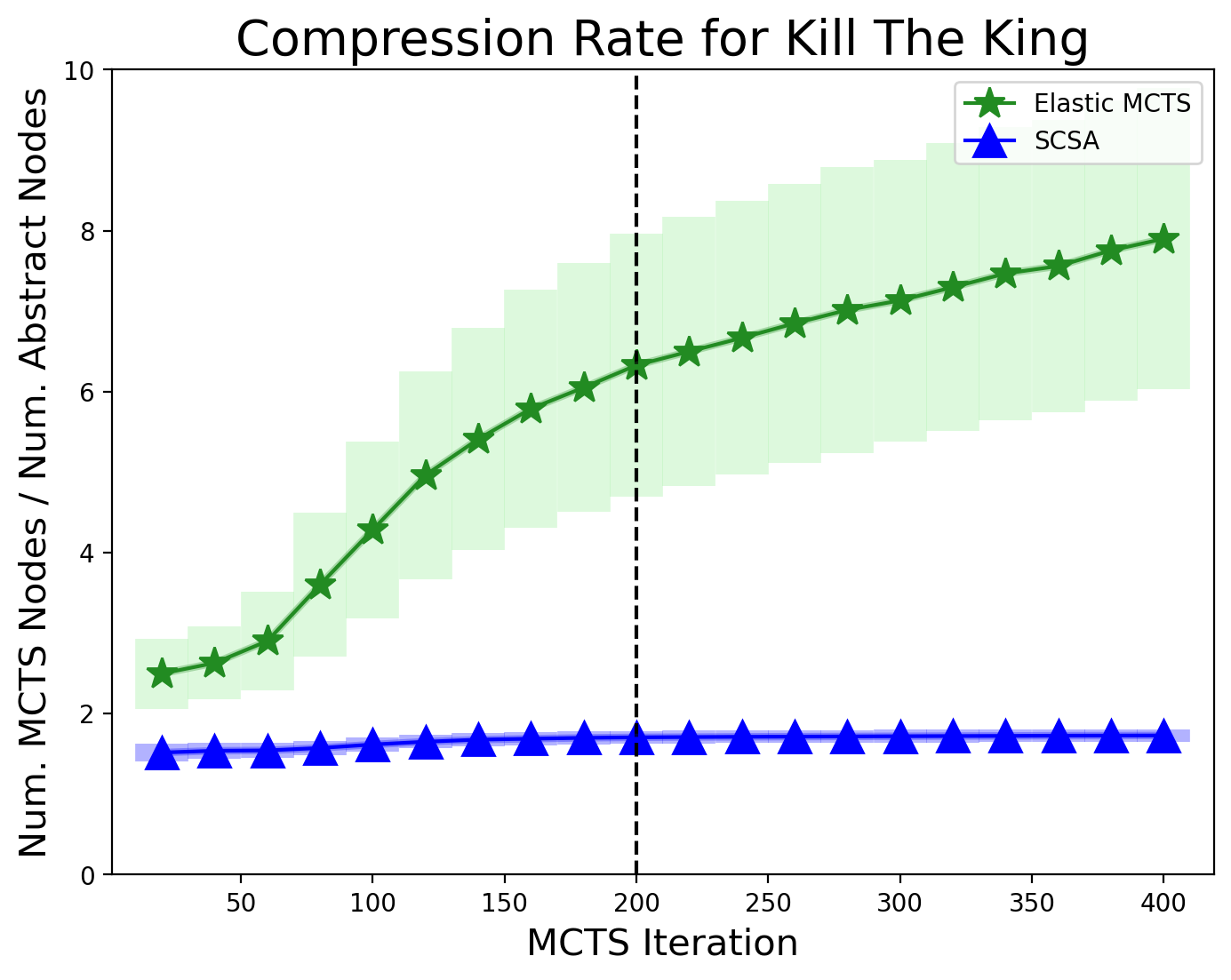}
        \label{fig:ktk_cr}
	}\hfill
    \subfloat[\normalsize{Push Them All (\textit{PTA})}]{
		\centering
        \includegraphics[width=0.6\columnwidth,trim=0 0cm 0cm 0.92cm, clip]{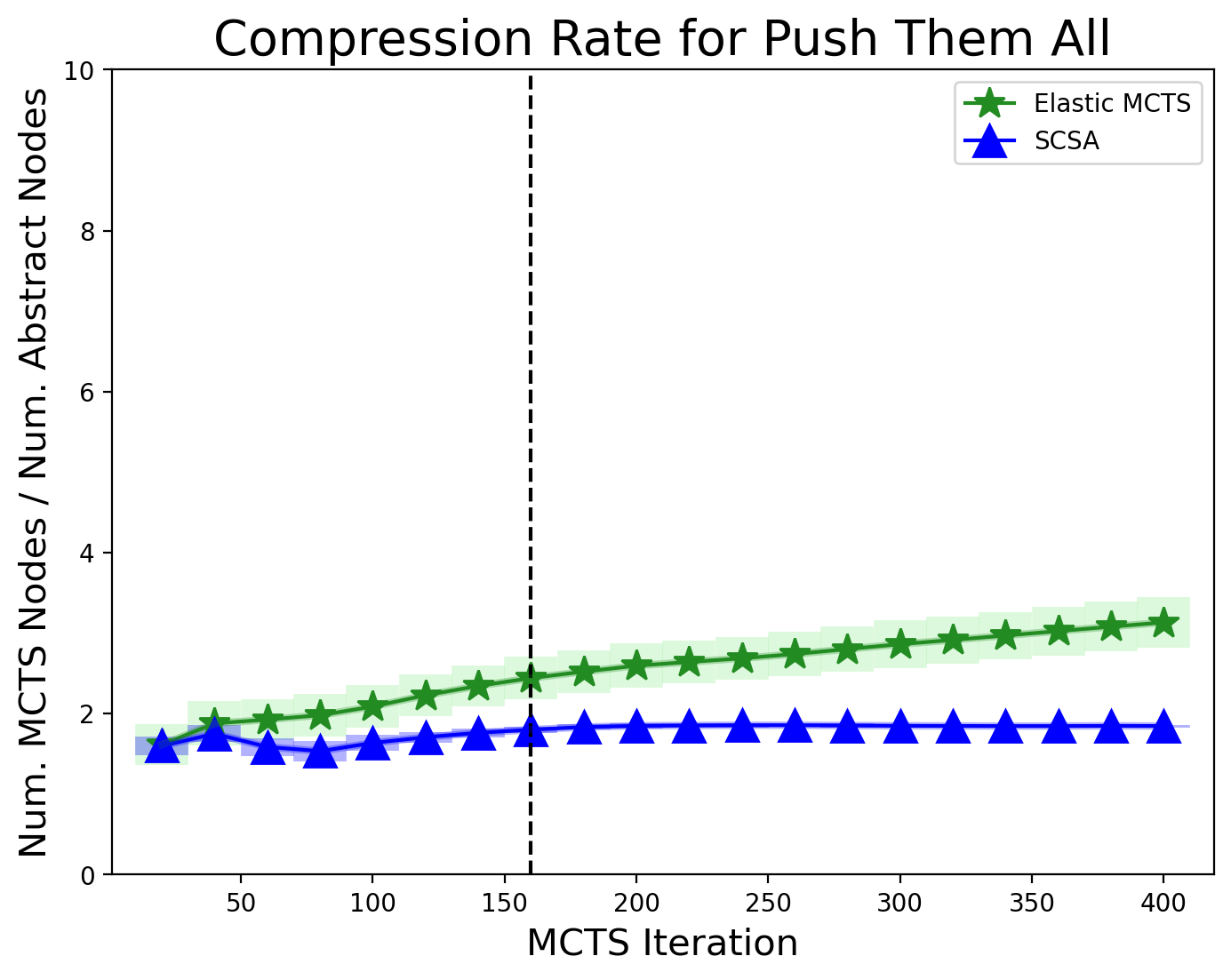}
        \label{fig:pta_cr}
	}\hfill
    \subfloat[\normalsize{Two Kingdoms (\textit{TK})}]{
		\centering
        \includegraphics[width=0.6\columnwidth,trim=0 0cm 0 0.92cm, clip]{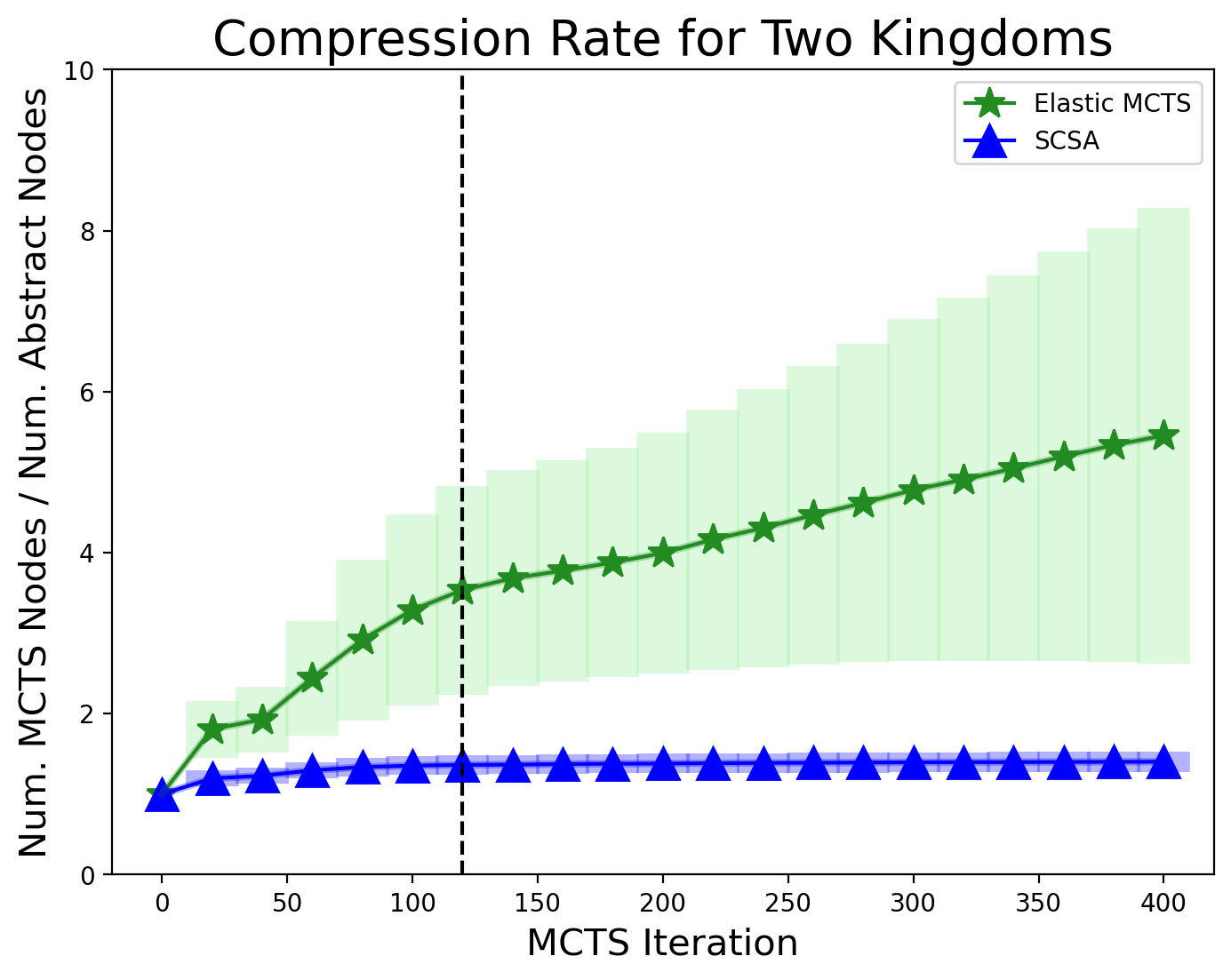}
        \label{fig:tk_cr}
	}
	\caption{Compression rates for each tested game including standard errors from 10 game plays. The vertical line indicates the iteration when the abstraction is split into the original node for the \eMCTSu.}
\end{figure*}

\subsection{Compression Rate}
To compare the influence of different state abstractions on tree size, we visualize compression rate at different MCTS iterations (see Figure~\ref{fig:ktk_cr}-\ref{fig:tk_cr}). The compression rate is defined as the number of tree nodes dividing the number of abstract nodes. We can see that the \textit{SIZE\_LIMIT} has constrained the compression rate by limiting the maximal size of each abstract node.
The overall compression rates of Elastic MCTS$_u$ are higher than SCSA and differ in different games. The vertical lines in Figure~\ref{fig:ktk_cr}-\ref{fig:tk_cr} indicate the iteration when Elastic MCTS$_u$ drops state abstraction.

We observe from the plots that Elastic MCTS$_u$ without \text{early stop} can obtain a higher compression rate but this degrades the performance (See~\citep{xu2023}). The SCSA agent outperforms Elastic MCTS$_u$ in two of the three games and it has lower compression rates. These observations reveal a trade-off between the abstracted tree size and the agent performance.

\section{Conclusion and Future Work}
Automatic state abstraction has recently been applied to MCTS to address large search spaces in strategy game-playing. However, the lack of data results in state abstraction of unstable quality. We propose the novel SCSA to control the abstraction quality. Compared to the previous \textit{early stop} approach, our method has a much smaller range for its hyperparameter. The empirical results on $3$ strategy games of different complexity present the effectiveness of SCSA on strategy game playing.

The SCSA outperforms baselines in two games but not in the complex $TK$ game, indicating a potential shortcoming of scalability. A possible solution is to combine state abstraction with pruning. We plan to further investigate the scalability of the SCSA agent in our future work. 

\textbf{Limitation} We analyze the tree size under different state abstraction size constraints, revealing a trade-off between memory usage and agent performance.

\section{Acknowledgement}
Work supported by UK EPSRC grant EP/T008962/1.

\bibliographystyle{unsrt}
\bibliography{cog24}

\end{document}